\pdfoutput=1
\documentclass[11pt]{article}
\usepackage[preprint]{acl}
\usepackage{times}
\usepackage{latexsym}
\usepackage[T1]{fontenc}
\usepackage[utf8]{inputenc}
\usepackage{microtype}
\usepackage{inconsolata}
\usepackage{graphicx}
\usepackage{multirow}
\usepackage{booktabs}
\usepackage{amsmath}
\usepackage{algorithm}
\usepackage{algorithmic}
\usepackage{amsfonts}
\usepackage{subcaption}

\usepackage{listings}
\usepackage{enumitem}
\usepackage{soul}
\definecolor{darkgreen}{rgb}{0, 0.5001960, 0}
\definecolor{darkred}{rgb}{0.8, 0, 0}

\title{ReasonGraph: Visualisation of Reasoning Paths}

\author{Zongqian Li \\
  University of Cambridge \\
  \texttt{zl510@cam.ac.uk} \\\And
  Ehsan Shareghi \\
  Monash University \\
  \texttt{ehsan.shareghi@monash.edu} \\\And
  Nigel Collier \\
  University of Cambridge \\
  \texttt{nhc30@cam.ac.uk}}

\begin{document}
\maketitle
\begin{abstract}
Large Language Models (LLMs) reasoning processes are challenging to analyze due to their complexity and the lack of organized visualization tools. We present \textbf{ReasonGraph}, a web-based platform for visualizing and analyzing LLM reasoning processes. It supports both sequential and tree-based reasoning methods while integrating with major LLM providers and over fifty state-of-the-art models. ReasonGraph incorporates an intuitive UI with meta reasoning method selection, configurable visualization parameters, and a modular framework that facilitates efficient extension. Our evaluation shows high parsing reliability, efficient processing, and strong usability across various downstream applications. By providing a unified visualization framework, ReasonGraph reduces cognitive load in analyzing complex reasoning paths, improves error detection in logical processes, and enables more effective development of LLM-based applications. The platform is open-source, promoting accessibility and reproducibility in LLM reasoning analysis.
\footnote{\url{https://github.com/ZongqianLi/ReasonGraph}}
\end{abstract}

\section{Introduction}
\label{Introduction}

Reasoning capabilities have become a cornerstone of Large Language Models (LLMs), yet analyzing these complex processes remains a challenge \citep{huang-chang-2023-towards}. While LLMs can generate detailed text reasoning output, the lack of process visualization creates barriers to understanding, evaluation, and improvement \citep{qiao-etal-2023-reasoning}. This limitation carries three key implications: (1) Cognitive Load: Without visual graph, users face increased difficulty in parsing complex reasoning paths, comparing alternative approaches, and identifying the distinctive characteristics of different reasoning methods \citep{li2024500xcompressorgeneralizedpromptcompression, li2024promptcompressionlargelanguage}; (2) Error Detection: Logical fallacies, circular reasoning, and missing steps remain obscured in lengthy text outputs, impeding effective identification and correction of reasoning flaws; and (3) Downstream Applications: The absence of standardized visualization frameworks restricts the development of logical expression frameworks and productivity tools that could improve and enrich LLM applications. These challenges highlight the essential need for unified visualization solutions that can illustrate diverse reasoning methodologies across the growing ecosystem of LLM providers and models.

To solve these challenges, we present \textbf{ReasonGraph}, a web-based platform for visualizing and analyzing LLM reasoning processes. The platform implements six mainstream sequential and tree-based reasoning methods and integrates with major LLM providers including Anthropic, OpenAI, Google, and Together.AI, supporting over 50 state-of-the-art models. ReasonGraph provides user-friendly UI design with intuitive components, real-time visualization of reasoning paths using Mermaid diagrams, meta reasoning method selection, and configurable parameter settings. The platform's modular framework enables easy integration of new reasoning methods and models while maintaining consistent visualization and analysis capabilities.

Our work makes three main \textbf{contributions}:

\vspace{-7pt}

\begin{itemize}[left=0pt, itemsep=0pt, parsep=0pt]
   \item \textbf{Unified Visualization Platform:} The first web-based platform that enables real-time graphical rendering and analysis of LLM reasoning processes, facilitating comparative analysis across different methods.
   
   \item \textbf{Modular and Extensible Design:} A flexible framework with modular components for easy reasoning methods and model integrations through standardized APIs.
   
   \item \textbf{Multi-domain Applications:} An open-source platform that bridges academia, education, and development needs, facilitating accessibility and reproducibility in LLM reasoning analysis.
\end{itemize}

The \textbf{paper structure} is organized as follows: Section \ref{Related Work} reviews related work in LLM reasoning methods and visualization approaches. We then detail our UI design principles and layout organization in Section \ref{UI Design}, followed by a presentation of our visualization methodology for both tree-based and sequential reasoning processes in Section \ref{Reasoning Visualisation}. Section \ref{Framework} elaborates the platform's modular framework and implementation details, while Section \ref{Applications} demonstrates the platform's versatility through various applications in academia, education, and development. After evaluating the platform's performance and usability in Section \ref{Evaluation}, we conclude in Section \ref{Conclusions} with a discussion of future directions.

\section{Related Work}
\label{Related Work}

LLM \textbf{reasoning methods} can be categorized into sequential reasoning and tree-based search approaches. Sequential reasoning, pioneered by Chain-of-Thought prompting \citep{wei2022chain}, demonstrates step-by-step problem decomposition and has been improved through multiple variants: Self-consistency \citep{wang2023selfconsistency} employs majority voting across multiple reasoning chains, Least-to-Most \citep{zhou2023leasttomost} decomposes complex problems into ordered sub-questions, and Self-refine \citep{madaan2023selfrefine} implements iterative reasoning refinement. Complementarily, tree-based approaches offer broader solution space exploration: Tree-of-Thoughts \citep{yao2023tree} enables state-based branching for parallel path exploration, while Beam Search reasoning \citep{freitag-al-onaizan-2017-beam} comprehensively evaluates solution paths based on scoring mechanisms, enabling efficient exploration of the reasoning space while maintaining solution diversity.

\textbf{Visualization approaches} for LLM reasoning processes have developed along two main directions: model behavior analysis and reasoning process illustration. In model behavior analysis, tools such as BertViz \citep{vig2019visualizingattentiontransformerbasedlanguage} and Transformers Interpret \citep{transformers-interpret}, while providing detailed visualizations of attention mechanisms and internal states, are limited to low-level model behaviors without showing higher-level reasoning characteristics. For reasoning process illustration, frameworks such as LangGraph \citep{langgraph} in LangChain \citep{langchain} offer only basic flow visualization for LLMs without supporting diverse reasoning methodologies, while general-purpose tools such as Graphviz \citep{graphviz} and Mermaid \citep{mermaid}, though flexible in graph creation, lack adaptions for LLM reasoning analysis. ReasonGraph introduced in this paper addresses these limitations by providing an open-source platform that supports multiple reasoning methods and various models, offers real-time visualization updates, and enables comprehensive analysis of reasoning processes.

\section{UI Design}
\label{UI Design}

\begin{figure*}[ht!]
    \centering
    \includegraphics[width=0.99\textwidth]{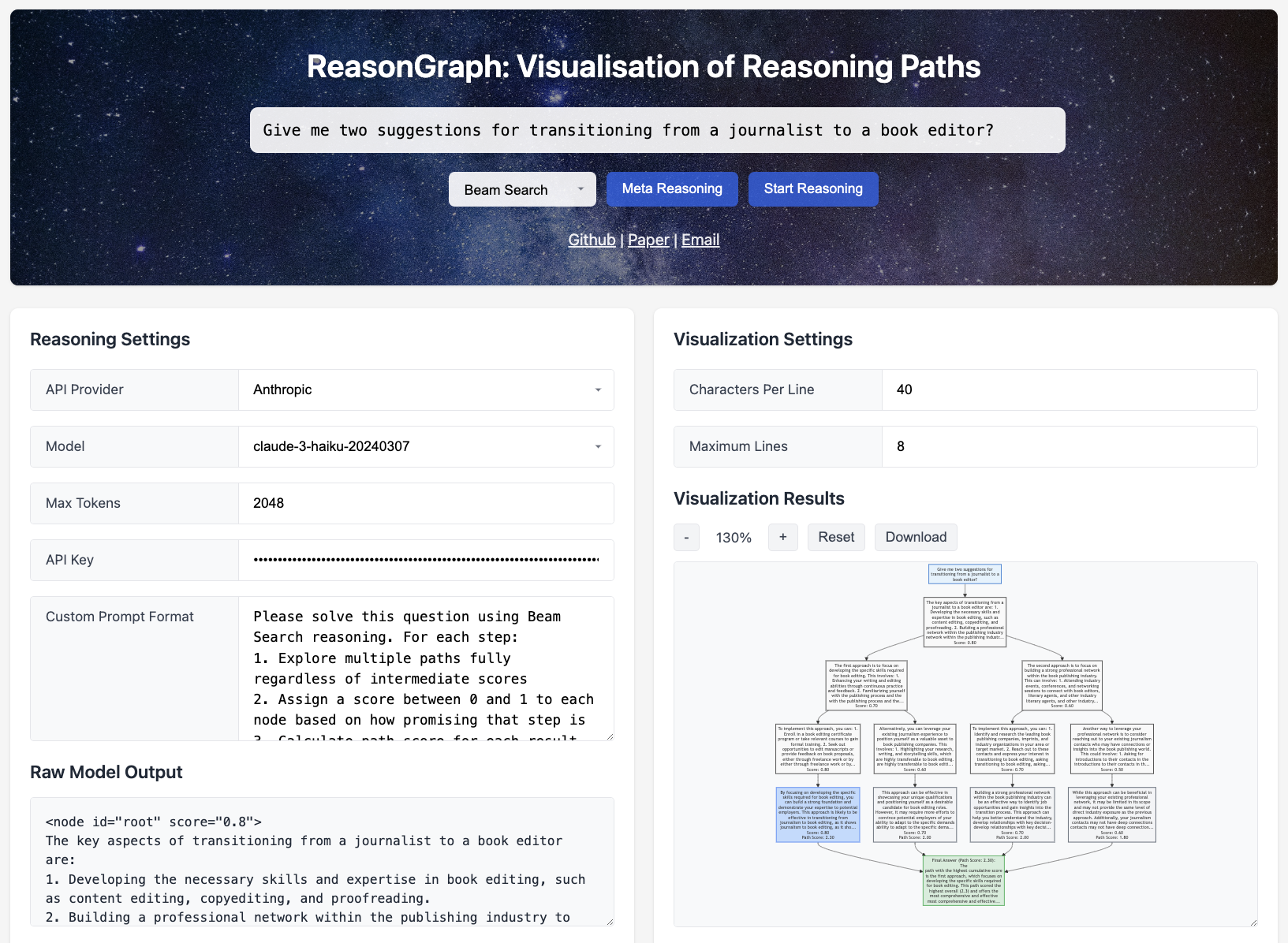}
    \caption{The ReasonGraph \textbf{UI} with a query input header and dual-panel layout: the left panel shows Reasoning Settings for configuring model API parameters along with Raw Model Output displaying the unprocessed text response from the LLM. The right panel contains Visualization Settings for adjusting diagram parameters and Visualization Results showing a flowchart illustration of the reasoning process.}
    \label{UI}
\end{figure*}

The \textbf{UI} of ReasonGraph shown in Figure \ref{UI} employs a two-column layout with a prominent header section for reasoning process visualization. The header section contains a central query input field, a reasoning method dropdown menu for manual method selection (e.g., Chain-of-Thoughts, Beam Search), and two buttons: "Meta Reasoning" for meta reasoning method selection by the model, and "Start Reasoning" for using the currently selected method. The main UI consists of two panels: the left panel combines Reasoning Settings for API configuration and model selection with Raw Model Output that displays the model's original text response, while the right panel pairs Visualization Settings for diagram parameters with Visualization Results that renders a graph illustration of the reasoning process, complete with zoom, reset, and export.

The UI design includes four fundamental product \textbf{design principles}: (1) Functional completeness: incorporating comprehensive model options, reasoning methods, and parameter settings to support diverse analytical needs; (2) Organized layout: maintaining a clear visual organization with the query input prominently positioned in the header, followed by parallel columns for text and graph outputs; (3) Universal usability: offering both manual method selection and model-recommended approaches to accommodate users' decision-making preferences; (4) Visual aesthetics: utilizing an elegant header background and alternating gray-white sections to create an organized appearance while preserving functional clarity \citep{D4DD00307A}. 

\section{Reasoning Visualisation}
\label{Reasoning Visualisation}

\begin{figure*}[ht!]
    \centering
    \includegraphics[width=0.99\textwidth]{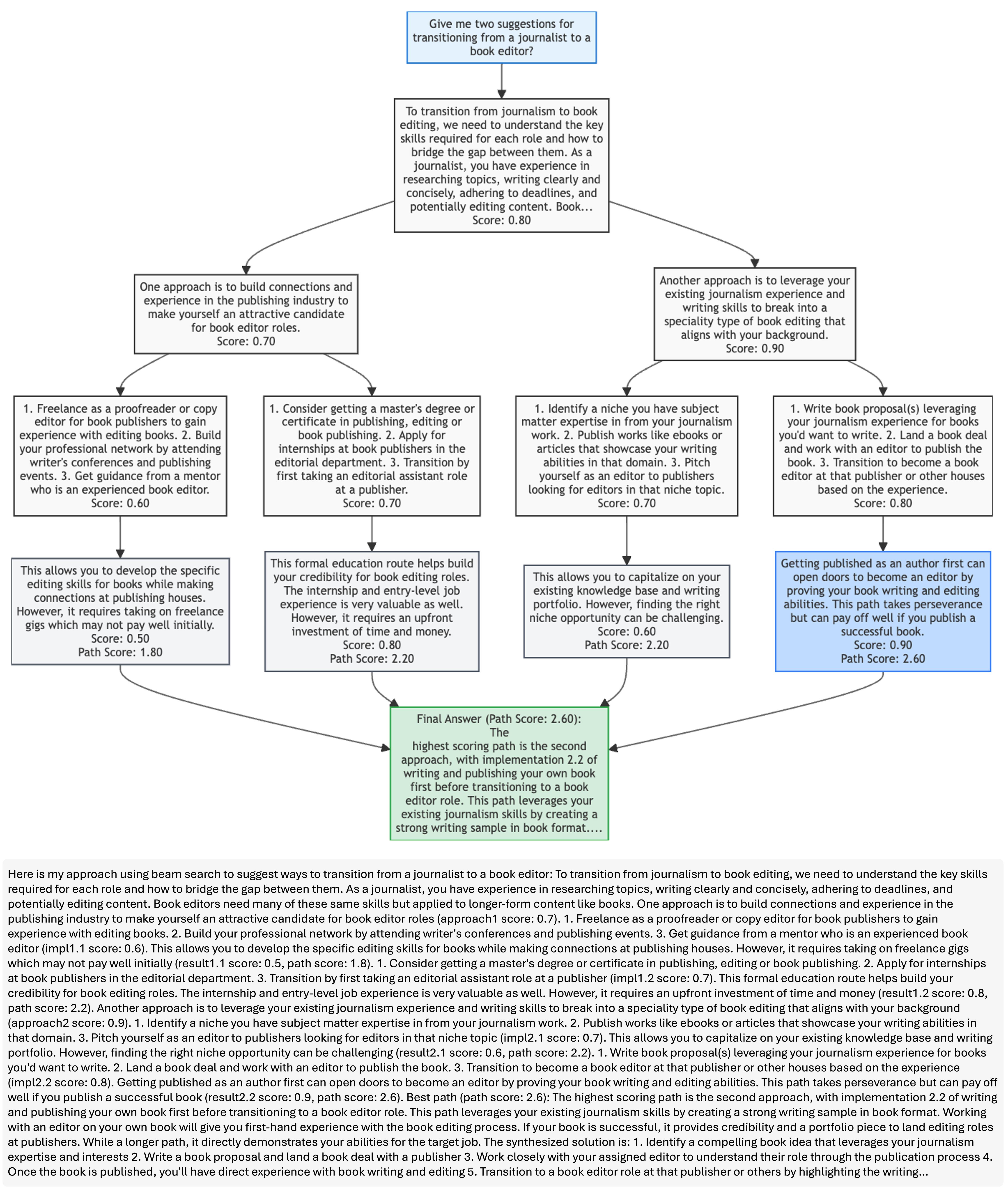}
    \caption{Comparison between plain text (bottom) and organized \textbf{tree visualization} (top) for the same reasoning process using beam search method. The blue box is the initial question, the darker blue box highlights the selected reasoning path, and the final solution is shown in a green box.}
    \label{tree_example}
\end{figure*}

Figure \ref{tree_example} illustrates the contrast between traditional text output and our organized visualization for a \textbf{tree-based search method}, beam search. In its visualization, each node denotes a reasoning step with a designated score, and each level maintains a consistent branching width, allowing for comprehensive exploration of solution spaces. The cumulative path scores guide the final solution selection, with the optimal path determined by the highest total score across all levels. While this method shares similarities with Tree-of-Thoughts visualization, the latter differs in its variable branching number and focus on state-space exploration rather than score-based progression. The visualization approach demonstrates clear advantages over raw text output: it provides immediate layout comprehension, enables quick identification of decision points, and facilitates direct comparison of alternative reasoning paths. The graphical illustration also makes the scoring mechanism and path selection process more clear, allowing users to trace the development of reasoning and understand the basis for the final solution.

\begin{figure*}[ht!]
    \centering
    \includegraphics[width=0.99\textwidth]{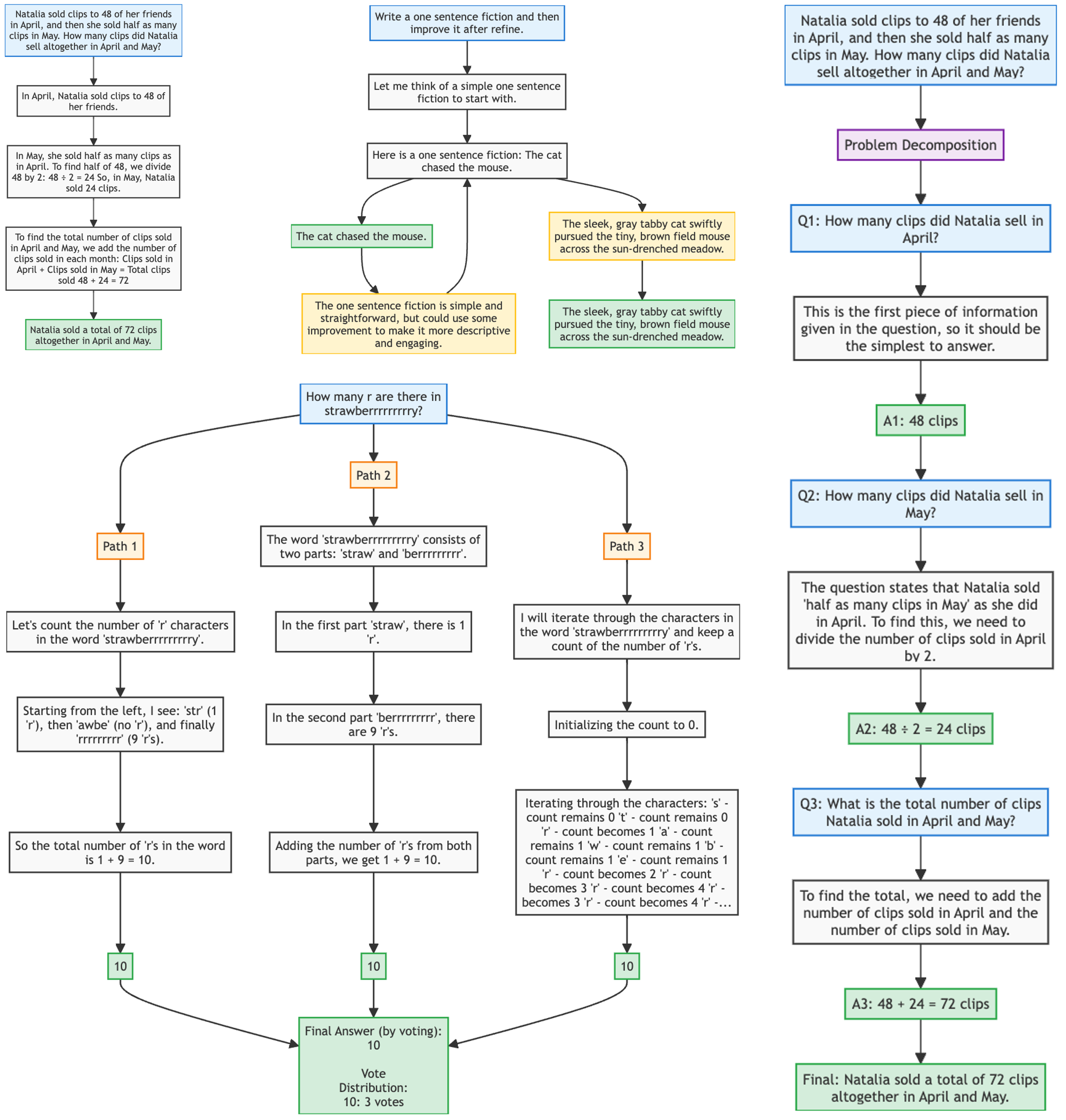}
    \caption{Visualization examples of four \textbf{sequential reasoning methods}: Chain-of-Thoughts (top-left), Self-refine (top-center), Least-to-Most (top-right), and Self-consistency (bottom-left). In Self-refine, yellow boxes indicate reflection and improvement steps; in Least-to-Most, light blue boxes are original and decomposed questions while green boxes show intermediate and final answers.}
    \label{sequence_example}
\end{figure*}

\textbf{Sequential reasoning processes} are visualized through directed graph layouts, as demonstrated in Figure \ref{sequence_example}. The visualization illuminates the step-by-step progression of different reasoning methods: Chain-of-Thoughts (top-left) displays a linear sequence of deductive steps leading to a final solution; Self-refine (top-center) shows the initial attempt followed by iterative improvements with refinement steps; Least-to-Most (top-right) demonstrates problem decomposition into simpler sub-questions with progressive solution building; and Self-consistency (bottom-left) illustrates multiple parallel reasoning paths converging to a final answer through majority voting. Each method's unique characteristics are exhibited through distinct visual layouts: linear chains for Chain-of-Thoughts, refinement loops for self-refine, leveled decomposition for Least-to-Most, and converging paths for self-consistency reasoning.

\begin{figure*}[ht!]
    \centering
    \includegraphics[width=0.95\textwidth]{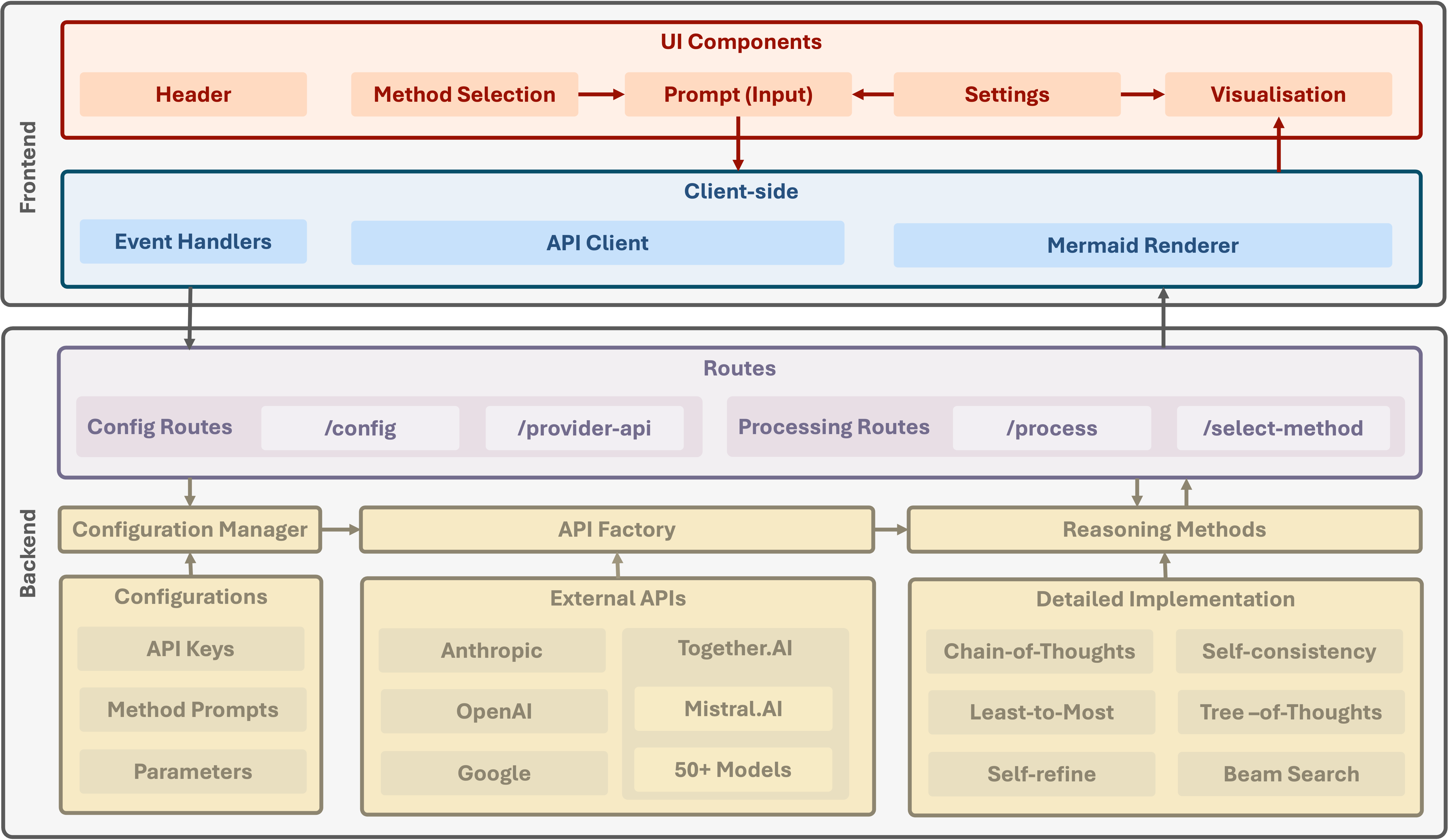}
    \caption{The \textbf{framework} of ReasonGraph, consisting of four main layers: UI Components for user involvement, Client-side for frontend processing, RESTful Routes for API bridge, and a modular backend comprising Configuration Manager, API Factory for LLM integration, and Reasoning Methods implementation.}
    \label{cover_figure}
\end{figure*}

\section{Framework}
\label{Framework}

ReasonGraph employs a modular framework that facilitates extensible reasoning visualization through separation of components.

The \textbf{frontend} tier encapsulates visualization logic and user participation handling. The layer implements an asynchronous event handling module, where user involvements with method selection and parameter configuration trigger corresponding state updates. The visualization module leverages Mermaid.js for dynamic graph rendering, with configurable parameters for node density and layout optimization, enabling real-time updates of reasoning process visualizations.

The \textbf{backend} framework is organized around three core modules implemented in Flask: a Configuration Manager for state update, an API Factory for LLM integration, and a Reasoning Methods module for reasoning approach encapsulation. The backend employs a RESTful API layer that ensures component connectivity and robust error handling, making it suitable for both academia and production scenarios.

The framework implements \textbf{modularity} at both API and reasoning method levels. The API Factory provides a unified API for multiple LLM providers through the BaseAPI class, while each reasoning method is encapsulated as an independent module with standardized API for parsing and visualization. This design enables dynamic switching between providers and reasoning methods, facilitating platform extension without framework modifications and ensuring adaptability to LLM capabilities.

\section{Applications}
\label{Applications}

ReasonGraph serves diverse use cases across academia, education, and development domains. For academic applications, it enables thorough analysis of LLM reasoning processes, facilitating comparative studies of different reasoning methods and evaluation of model capabilities across various tasks. In educational contexts, the platform serves as an efficient tool for teaching logical reasoning principles and demonstrating LLM decision-making processes, while helping students understand the strengths and limitations of different reasoning approaches. For development purposes, ReasonGraph helps prompt engineering optimization by visualizing how different prompts influence reasoning paths, supports refinement of LLM applications through clear visualization of reasoning paths, and assists in selecting optimal reasoning methods for specific task types.

\section{Evaluation}
\label{Evaluation}

The evaluation of ReasonGraph demonstrates our platform's robustness in three key aspects: (1) parsing reliability, where our rule-based XML parsing approach achieves nearly 100\% accuracy in extracting and visualizing reasoning paths from properly formatted LLM outputs; (2) processing efficiency, where the Mermaid-based visualization generation time is negligible compared to the LLM's reasoning time, maintaining consistent performance across all six reasoning methods; and (3) platform usability, where preliminary feedback from open-source platform users indicates that approximately 90\% of users successfully used the platform without assistance, though these metrics continue to change as our user base expands and the platform undergoes regular updates. The overall platform performance depends on the LLM's ability to follow the specified instructions, with guaranteed parsing accuracy for well-formatted responses.

\section{Conclusions}
\label{Conclusions}

This paper introduces \textbf{ReasonGraph}, a web-based platform that enables visualization and analysis of LLM reasoning processes across six mainstream methods and over 50 models. Through its modular framework and real-time visualization capabilities, the platform achieves high usability across diverse applications in academia, education, and development, improving the understanding and application of LLM reasoning processes.

\textbf{Future work} will pursue four key directions. First, we will leverage the open-source community to integrate additional reasoning methods and expand model API support. Second, we plan to develop the platform based on community feedback and user suggestions, improving platform usability and functionality. Third, we will continue exploring downstream applications such as reasoning evaluation, educational tutorials, and prompting tools. Finally, we aim to implement editable nodes in the visualization flowcharts, enabling direct modification of reasoning processes through the graph workspace.

\section*{Limitations}

The current development of ReasonGraph has been primarily done by individual efforts, which naturally limits its scope. A broader open-source community effort is needed to improve the platform's performance, identify potential issues in usage, and collaboratively improve the platform's overall function completeness.

\section*{Ethics Statement}
No ethical concerns are present.

\section*{Availability Statement}
The codes related to this paper have been uploaded to \url{https://github.com/ZongqianLi/ReasonGraph}.

\bibliography{custom}

\end{document}